\documentclass{article}

\PassOptionsToPackage{square,numbers,sort&compress}{natbib}

\usepackage[main, final]{neurips_2026}

\usepackage[utf8]{inputenc} 
\usepackage[T1]{fontenc}    
\usepackage[colorlinks,linktoc=all]{hyperref}       
\usepackage[dvipsnames]{xcolor}
\hypersetup{linkcolor=MidnightBlue,citecolor=MidnightBlue,urlcolor=MidnightBlue}
\usepackage{url}            
\usepackage{booktabs}       
\usepackage{amsfonts}       
\usepackage{nicefrac}       
\usepackage{microtype}      
\usepackage{xcolor}         
\usepackage{graphicx}
\usepackage{subcaption}
\usepackage{amsmath}
\usepackage{bm}
\usepackage{amssymb}
\usepackage{mathtools}
\usepackage{amsthm}
\usepackage{algorithm}
\usepackage[noend]{algpseudocode}
\usepackage{multicol}
\usepackage{multirow}
\usepackage{wrapfig}

\newcommand{\diff}{\mathop{}\!\mathrm{d}}

\theoremstyle{plain}
\newtheorem{theorem}{Theorem}[section]
\newtheorem{proposition}[theorem]{Proposition}

\theoremstyle{definition}

\theoremstyle{remark}

\title{ESS-Flow: Training-free guidance as Bayesian inference in source space}

\author{
Adhithyan Kalaivanan$^1$ \quad Zheng Zhao$^1$ \quad Jens Sjölund$^2$ \quad Fredrik Lindsten$^{1}$ \\
$^1$Linköping University, Sweden \quad $^2$Uppsala University, Sweden\\
\texttt{\{adhithyan.kalaivanan,zheng.zhao,fredrik.lindsten\}@liu.se}\\
\texttt{jens.sjolund@it.uu.se}\\
}

\begin{document}

\maketitle

\begin{abstract}
Guiding pretrained flow-based generative models for conditional generation or to produce samples with desired target properties enables solving diverse tasks without retraining on paired data.
We present ESS-Flow, a gradient-free method that leverages the Gaussian prior of the source distribution in flow-based models to perform Bayesian inference directly in the source space using elliptical slice sampling.
ESS-Flow only requires forward passes through the generative model and observation process, making it applicable even when gradients are unavailable, such as with non-differentiable simulation-based observations or quantization in the generation process.
We demonstrate its effectiveness on scientific applications including generating materials with target properties, protein backbone structures with improved designability, and synthesizable, drug-like molecules. The associated code is available at \href{https://github.com/Adhithyan8/essflow}{https://github.com/Adhithyan8/essflow}.
\end{abstract}

\section{Introduction}
In generative modeling, we are given data samples and aim to construct a sampler that approximates the data distribution.
Diffusion models \citep{ho2020denoising, song2020score} and continuous normalizing flows \citep{lipman2022flow, liu2022flow, albergo2023stochastic} achieve this by transporting samples from a simple source distribution to the data distribution.
Beyond unconditional sampling, we often need to generate samples with specific properties or to match observed data in the context of inverse problems.
If sufficient paired data are available, we can train specialized conditional generative models \citep{dhariwal2021diffusion, rombach2022high, miller2024flowmm} for each task.

Training-free conditional generation methods offer a flexible alternative by reusing pretrained models.
Instead of paired training data, these methods typically use a likelihood (or more generally, a potential) function to measure how well a sample matches the desired target.
Guidance-based methods modify the generative process using the score of the likelihood \citep{song2020score}, though this often requires approximations.
Optimization-based methods minimize the negative log-likelihood with gradient descent, using the generative model as an explicit regularizer through noising-denoising procedures \citep{levy2024solving, martin2025pnp} or an implicit regularizer that constrains the gradient flow to the data manifold \citep{ben2024d}.
Alternatively, sampling-based approaches formulate controlled generation as Bayesian inference with the pretrained model as the prior. However, existing methods require likelihood gradients \citep{chung2023diffusion, wu2023practical, janati2024divide, zhang2025improving, janati2025mixture} or are restricted to linear Gaussian observations \citep{cardoso2024monte, kelvinius2025solving}.

These methods all rely on differentiating through the observation process, and fail when such gradients are unavailable.
This limits their applicability in scientific domains where the observation or reward function is often computed by non-differentiable external simulators or involves discrete operations that are not amenable to automatic differentiation \citep[e.g.][]{alhossary2015fast, alford2017rosetta}.
Gradient-free alternatives for sampling in the data space could be considered, but these require designing proposal distributions over a complex, high-dimensional data manifold and careful tuning of step sizes.

\begin{wrapfigure}{r}{0.5\textwidth}
\centering
\vspace{-5.5mm}
\includegraphics[trim={0 3mm 0 0}, clip, width=\linewidth]{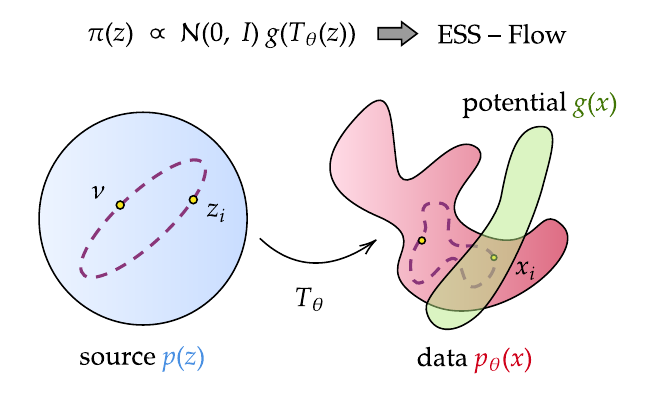}
\caption{Illustration of ESS-Flow. We sample the target distribution $\pi(z)$ in the source space of flow-based models, which typically has a Gaussian prior $p(z)$. This allows gradient-free sampling with elliptical slice sampling. At stationarity, the transformed ellipse passes through regions of high potential in the data space by construction.}
\label{fig:essflow}
\vspace{-3.5mm}
\end{wrapfigure}

Our key insight is that many pretrained diffusion and flow-based models can be viewed as continuous normalizing flows with Gaussian source distributions, enabling Bayesian inference in the source space instead of the complex data space.
The Gaussian prior allows us to use elliptical slice sampling \citep[ESS,][]{murray2010elliptical}, a Markov chain Monte Carlo (MCMC) method that generates proposals along ellipses defined by the current state and a draw from the prior.
ESS requires no hyperparameter tuning and provides adaptive step sizes through a shrinkage mechanism.
This results in ESS-Flow, a gradient-free controlled generation method that approximates the target distribution in the source space of flow-based models, as illustrated in Figure \ref{fig:essflow}.

ESS-Flow can be used with any generative prior, including those trained with Riemannian flow matching on manifold-valued data, as long as the source distribution is Gaussian and the transport map is deterministic.
Furthermore, unlike guidance-based methods, ESS-Flow preserves the pretrained vector field. 
As discussed by \citet{wang2025source}, this has the benefit that if the prior vector field is optimized for fast generation, for example when trained with minibatch-OT coupling \citep{pooladian2023multisample, tong2024improving}, this property is retained.

There exists a plethora of methods \citep{choi2021ilvr, kawar2022denoising, song2023pseudoinverse, chung2023diffusion, pokle2024training} designed for image restoration problems, such as inpainting, denoising, and superresolution. Our method addresses a complementary class of problems with important use-cases in scientific applications. There are at least two key differences between these application domains.
 
First, many image restoration problems have differentiable and highly informative likelihoods.
For instance, in inpainting, the exact equality constraints for observed pixels concentrate the posterior on a low-dimensional manifold in the data space.
In such cases, leveraging the available likelihood gradients to steer samples toward this manifold is highly beneficial.
In scientific applications, however, we often condition on weakly informative (highly noisy) measurements or target specific properties where the design space remains flexible and non-degenerate.
This makes the sampling more amenable to gradient-free methods.
Furthermore, non-differentiable likelihoods are more common in scientific applications, ruling out gradient-based methods.
We demonstrate these problem types in Section \ref{sec:experiments}, confirming the effectiveness of ESS-Flow in such cases.
 
Second, for applications in computer vision, perceptual realism in the reconstruction is often the primary objective. 
This is in stark contrast with many scientific applications where we need sufficient diversity for exploratory analysis and/or reliable uncertainty quantification for hypothesis testing.
In such applications, capitalizing on the statistical consistency of well-established Monte Carlo methods, as ESS-Flow does, ensures statistical rigor in the analysis. 

We summarize our main contributions as:
\begin{itemize}
\item We present ESS-Flow, a training-free and asymptotically exact sampling method for controlled generation that works with non-differentiable observations and discrete operations.
\item We demonstrate ESS-Flow on generating materials with target properties, achieving lower mean absolute errors and higher sample quality compared to gradient-based methods. We additionally apply ESS-Flow to protein backbone design and molecule generation with non-differentiable rewards, improving designability and drug-likeness, respectively.
\end{itemize}

\section{Preliminaries}

\paragraph{Flow-based generative models.}
Consider a generative model that transports samples $x_0$ from a simple source distribution at $t=0$ to samples $x_1$ from the data distribution at $t=1$ through a learned vector field $u_t^\theta(x)$ by solving the ordinary differential equation (ODE):
\begin{equation}
    x_1 = T_\theta(x_0) \coloneqq x_0 + \int_0^1 u_t^\theta(x_t) \diff t.
    \label{equ:flow-model}
\end{equation}

Both diffusion models with the probability flow ODE for generation \citep{song2020score} and models trained with the conditional flow matching objective \citep{lipman2022flow, albergo2023stochastic} can be viewed as instances of this model class.
We collectively refer to them as flow-based generative models with a transport map $T_\theta$ and consider the case where the
source distribution is Gaussian, i.e., the
model defines a continuous normalizing flow.
To simplify notation, we drop the time subscript $t$ and denote source samples $x_0$ by $z$ and data samples $x_1$ by $x$, i.e., the generative model prior is defined by $x = T_\theta(z)$ with $z\sim \mathcal{N}(0,I)$.

\paragraph{Controlled generation as Bayesian inference.}
In controlled generation, we seek samples from a target distribution $\pi(x) \propto g(x) p_\theta(x)$, where the normalization constant is assumed finite so that $\pi(x)$ is a probability distribution. Here,
$p_\theta(x)$ is the prior (unconditional) generative model and $g(x)$ is a nonnegative potential function guiding samples toward desired properties.
This formulation yields the posterior $p(x|y)$ when $g(x)$ is a likelihood $p(y|x)$ conditioned on observation $y$, or a tilted prior when $g(x)$ is a general reward function.
In our setting, the prior data distribution is given by the pretrained flow-based model of the form in equation~\eqref{equ:flow-model}.

\section{Related Work}
Many methods have been proposed for controlled generation with diffusion and flow-based models \citep[see surveys in, e.g.,][]{daras2024survey, chung2025diffusion, Zhao2024rsta}.
We focus on those that are training-free and applicable to ODE-based generative models of the form in equation~\eqref{equ:flow-model}.
This excludes directly training conditional generative models with paired data \citep{dhariwal2021diffusion, rombach2022high} and RL-based methods to finetune diffusion models \citep{venkatraman2024amortizing, zhaoadding}.

Methods such as DPS \citep{chung2023diffusion}, OT-ODE \citep{pokle2024training}, and FlowDPS \citep{kim2025flowdps} incorporate a guidance term into the transport map, obtaining target samples in a single pass through the modified generative process.
This involves the score of the likelihood, which is often approximated, and the sequential process cannot correct for errors that occur early in the generation.
Methods like PnP-Flow \citep{martin2025pnp}, DAPS \citep{zhang2025improving}, DDSMC \citep{kelvinius2025solving} and FK-Steering \citep{singhalgeneral} address these limitations by alternating between the noise space and the data space with progressively annealed noise.
PnP-Flow optimizes in the data space and regularizes through noising-denoising steps between gradient updates.
DAPS generalizes this framework for posterior sampling, and DDSMC extends DAPS using sequential Monte Carlo for linear Gaussian potentials.
These methods do not require differentiating the transport map and perform approximate posterior updates in the data space at each iteration of the generative procedure.
However, they require access to the noising process used during training.

Source space methods like D-Flow \citep{ben2024d} perform gradient-based optimization, relying on implicit regularization that constrains gradient flow to the data manifold.
\citet{purohit2025consistency} extend D-Flow to posterior sampling using Langevin Monte Carlo.
\citet{wang2025source} use Hamiltonian Monte Carlo in the source space, which was explored earlier by \citet{graham2017asymptotically} for variational autoencoders.
A key limitation of these methods is their reliance on gradients, requiring backpropagation through the ODE solver.
Outsourced Diffusion Sampling \citep{venkatraman2025outsourced}, which operates in the source space, is gradient-free and can be viewed as variational inference, where the sampler is amortized through additional training.
In concurrent work, \citet{wang2026testtime} present gradient-free MCMC in the source space by combining preconditioned Crank--Nicolson (pCN) and parallel tempering over annealed rewards.
Elliptical slice sampling, used in our method, can be seen as a tuning-free, adaptive step-size variant of pCN.

\section{Method}
\label{sec:method}
Suppose that we have a generative model prior $p_\theta(x)$ that maps samples from a Gaussian source distribution $z \sim \mathcal{N}(0, I)$ to the data space with the transport map $x = T_\theta(z)$.
Given a potential function $g(x)$, such as the likelihood $p(y|x)$ for an observation $y$ or a general reward function, we seek samples from the target $\pi(x)$:
\begin{equation}
    \pi(x) \propto g(x) \, p_\theta(x)
    = g(x) \, \frac{p(z)}{\left|\det(J_{T_\theta}(z))\right|},
\end{equation}
where $z = T^{-1}_\theta(x)$ and $J_{T_\theta}(z)$ denotes the Jacobian of the transport map evaluated at $z$.

Sampling the target distribution $\pi(x)$ directly in the data space is challenging as exact density evaluation requires computing the Jacobian determinant, which is expensive for continuous normalizing flows. 
Moreover, gradient-free MCMC methods would require designing effective proposals on the complex, high-dimensional data manifold.
Instead, we can work in the source space, where the simple Gaussian prior 
enables efficient gradient-free sampling.

\subsection{ESS-Flow: gradient-free source space sampling}
With a change of variables, we can express the target distribution $\pi(z)$ in the source space:
\begin{equation}
    \pi(z) = \pi(x) \left|\det(J_{T_\theta}(z))\right| 
    \propto g(x) \, \frac{p(z)}{\left|\det(J_{T_\theta}(z))\right|} \left|\det(J_{T_\theta}(z))\right| 
    = g(T_\theta(z)) \, p(z).
\label{eq:sourceposterior}
\end{equation}
The Jacobian terms cancel when both the prior and the posterior are expressed in the source space, enabling straightforward point-wise density evaluation. 
We are interested in settings where gradients are unavailable, for instance, when the generative process involves quantization, or when evaluating the potential requires non-differentiable simulators.
This rules out gradient-based source space sampling methods such as Langevin Monte Carlo \citep{purohit2025consistency} or Hamiltonian Monte Carlo \citep{graham2017asymptotically, wang2025source}, which require computing $\nabla_z g(T_\theta(z)) = J_{T_\theta}(z)^\mathsf{T} \nabla_x g(x)$.
Moreover, even when approximate gradients can be computed, gradient-based methods in the source space can struggle when the data distribution has multiple disconnected modes.

\begin{wrapfigure}{r}{0.5\textwidth}
  \centering
  \vspace{-5mm}
  \begin{subfigure}[h]{0.24\textwidth}
    \centering
    \includegraphics[width=\textwidth]{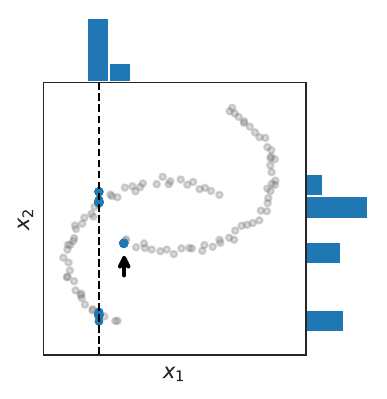}
    \caption{D-Flow samples}
    \label{fig:dflow}
  \end{subfigure}
  \hfill
  \begin{subfigure}[h]{0.24\textwidth}
      \centering
      \includegraphics[width=\textwidth]{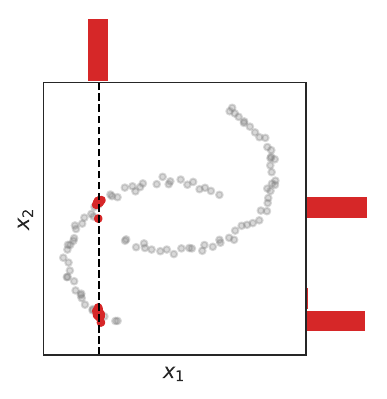}
      \caption{ESS-Flow samples}
      \label{fig:ess}
  \end{subfigure}
  \caption{Conditional generation on a 2D toy problem, targeting samples whose first coordinate $x_1$ is close to a value $c$ indicated by the dotted line. Prior samples are shown in gray with marginals along the border. Some D-Flow samples, which follow the source-space gradient, become trapped in disconnected manifold components (indicated by the arrow), while ESS-Flow avoids this problem.}
  \label{fig:moons}
  \vspace{-3mm}
\end{wrapfigure}

To illustrate this, consider D-Flow applied to a toy problem, where the prior $p_\theta(x)$ over $x = (x_1, x_2) \in \mathbb{R}^2$ consists of two interleaving half-circles, and the potential $g(x)$ favors samples whose first coordinate $x_1$ is close to a target value $c$. 
D-Flow maximizes $g(T_\theta(z))$ w.r.t.\ $z$ using gradient-based optimization while relying on implicit regularization.
A small gradient step $\delta_z \propto J_{T_\theta}(z)^\mathsf{T} \nabla_x g(x)$ in the source space corresponds to the step $\delta_x = J_{T_\theta}(z) J_{T_\theta}(z)^\mathsf{T} \nabla_x g(x)$ in the data space.
This projects gradients along the data manifold, resulting in a manifold-constrained gradient flow \citep{ben2024d}.
As a result, the source space gradient can vanish even when $\nabla_x g(x)$ is non-zero, and samples following the gradient flow can be trapped in local optima.
As seen in Figure~\ref{fig:moons}, even with a large learning rate that deviates significantly from a gradient flow, many D-Flow samples remain within the disconnected manifold component where they are initialized. This failure mode is evident in our material generation experiments in Section \ref{sec:material}.

We propose to use elliptical slice sampling~\citep[ESS,][]{murray2010elliptical} to sample from the target distribution in the source space.
This leverages the fact that the prior $p(z)$ is a multivariate Gaussian, while all the complexity of the transport map $T_\theta(z)$ is absorbed into the potential according to equation~\eqref{eq:sourceposterior}.
This is precisely the setup where ESS excels.
Our resulting approach, ESS-Flow, requires only point-wise evaluations of the generative model and involves no gradient or Jacobian computation.
Being an MCMC method, ESS-Flow is flexible with initialization, e.g., by sampling $z_0$ from the Gaussian prior, and then proceeding iteratively.
As illustrated in Figure \ref{fig:essflow}, a proposal is drawn from an ellipse in the source space defined by the current state $z_i$ and a random sample $\nu$ from the Gaussian prior.
Assuming that the transport map $T_\theta$ is continuous in $z$, it nonlinearly maps the ellipse to a continuous, connected curve in the data space that passes through the current state $x_i$.
The potential function $g(T_\theta(z))$ evaluated in the data space determines whether the proposal will be accepted.
Unlike Metropolis--Hastings, where a rejected proposal requires drawing a new ellipse, ESS adaptively shrinks the angular bracket around the current ellipse when a proposal is rejected.
This excludes regions containing the rejected proposal while maintaining support near the current state, from which a new proposal is then drawn.
This mechanism provides an adaptive step size that enables larger jumps without tuning.
As discussed by \citet{murray2010elliptical}, this guarantees that the procedure will terminate by accepting a proposal in finite time when the pullback potential function $g \circ T_\theta$ is continuous in a neighborhood around $z_i$.
Searching along the ellipse in the source space corresponds to gradient-free exploration of the target distribution in the data space, and the resulting Markov chain in the source space $\{z_i\}$ yields the corresponding Markov chain $\{x_i\}$ when passed through the transport map.
ESS-Flow has no hyperparameters to tune, and one MCMC step is summarized in Algorithm \ref{alg:essflow}.

ESS is known to perform well in high dimensions, with well-defined extensions to infinite-dimensional Hilbert spaces \citep{hasenpflug2025reversibility}.
\citet{murray2010elliptical} discuss the convergence of ESS by showing that the induced Markov kernel leaves the target distribution invariant.
A more detailed convergence and reversibility analysis is provided by \citet{Natarovskii2021} and \citet{hasenpflug2025reversibility}. 
For completeness, we state one of the main results here, adapted to our setting of flow-based models.

\begin{proposition}[Geometric convergence of ESS-Flow, Theorem~2.2 by \citet{Natarovskii2021}]
    Suppose that the pullback potential $z\mapsto g \circ T_\theta(z)$ is 
    bounded away from 0 and $\infty$ on compact sets and has regular tail behavior (details in \citet{Natarovskii2021}, Assumption~2.1), 
    then the ESS Markov chain, which we denote $\nu(\cdot, z)$, converges geometrically fast to the target measure $\pi$: 
    \begin{equation*}
        \lVert \nu^n(\cdot, z) - \pi \rVert_{\mathrm{TV}} \leq c \, (1 + \lVert z \rVert_2 ) \, \beta^n, \quad \forall z\in\mathbb{R}^{d},
    \end{equation*}
    where $\lVert \cdot \rVert_\mathrm{TV}$ stands for the total variation distance, $\nu^n$ is the $n$-th iteration of the chain, and $c>0$ and $0<\beta <1$ are constants. 
\end{proposition}

While the result does not characterize how $c$ and $\beta$ depend on dimensionality, \citet{Natarovskii2021} empirically find the statistical efficiency of ESS, in terms of the effective sample size, to be dimension-independent.
We also present numerical evaluations for the scaling of ESS-Flow with dimension in Appendix \ref{gmm} on a GMM prior with a tractable exact posterior.

\begin{algorithm}
\caption{ESS-Flow: one MCMC iteration}\label{alg:essflow}
\begin{algorithmic}
\Require Transport map $T_\theta$, potential $g$, current state and potential $(z, g(T_\theta(z)))$
\end{algorithmic}
\begin{algorithmic}
\State $\nu \sim \mathcal{N}(0, I)$ \Comment{Sample the variable which defines the ellipse}
\State $u \sim U(0, 1)$ \Comment{Sample the acceptance threshold for this step}
\State $\theta \sim U(0, 2\pi), \, [\theta_{l}, \theta_{u}] \leftarrow [\theta - 2\pi, \theta]$ \Comment{Initialize the angle and the bracket}
\While{true}
    \State $z' = z \cos \theta + \nu \sin \theta$ \Comment{Define the proposal along the ellipse}
    \If{$\log g(T_\theta(z')) > \log g(T_\theta(z)) + \log u$}
        \State \textbf{return} $(z', g(T_\theta(z')))$ \Comment{If the condition is met, accept the proposal}
    \Else
        \State \textbf{if} $\theta < 0$ \textbf{then} $\theta_{l} \gets \theta$ \textbf{else} $\theta_{u} \gets \theta$ \Comment{Else shrink the bracket to exclude the proposal}
        \State Resample $\theta \sim U(\theta_{l}, \theta_{u})$ \Comment{Sample new angle from the shrunk bracket}
    \EndIf
\EndWhile
\end{algorithmic}
\end{algorithm}

\subsection{Multi-fidelity sampling with ESS-Flow}
\label{sec:multifidelity}
Flow-based generative models are defined in continuous time according to equation~\eqref{equ:flow-model}, but in practice the transport map is solved numerically with finite discretization $\Delta$ rather than exactly.
This means that we have access to a class of approximate priors $\{p_\theta^\Delta(x) : \Delta \in (0,1)\}$ corresponding to different discretization levels, where smaller $\Delta$ gives rise to more accurate models. 

Our proposed MCMC-based sampler could be generalized to take advantage of such a multi-fidelity setup.
The high-level idea is to rely on coarser, and thus computationally cheaper, evaluations of the transport map for most of the sampling, while ensuring that the final samples nevertheless target a prespecified high-fidelity model.
This could be accomplished by delayed acceptance ESS \citep{bitterlich2025delayed}, or by methods resembling parallel tempering \citep{earl2005parallel} or simulated tempering \citep{marinari1992simulated} where the annealing temperature is replaced by the discretization level. 
A simpler approach, which we explore here as a proof of concept, is post-correction of the generated samples using importance weighting.

Let $T^\Delta_\theta(z)$ denote the transport map simulated with coarse discretization $\Delta$ and $T^{\delta}_\theta(z)$ with fine discretization $\delta \ll \Delta$.
The target distribution for the high-fidelity model from equation~\eqref{eq:sourceposterior} can be rewritten as:
\begin{equation}
    \pi^\delta(z) \propto g(T^{\delta}_\theta(z)) \, p(z) 
    = \frac{g(T^{\delta}_\theta(z))}{g(T^\Delta_\theta(z))} g(T^\Delta_\theta(z)) \, p(z)
    \propto w \, \pi^\Delta(z)
\label{eq:importanceweight}
\end{equation}

We sample from ESS-Flow targeting $\pi^\Delta(z)$ using the coarse transport map $T^\Delta_\theta$ and reweight samples with self-normalized importance weights $w_i \propto g(T^{\delta}_\theta(z_i))/g(T^\Delta_\theta(z_i))$.
The expensive high-fidelity transport map is evaluated only on the final MCMC samples, not during ESS itself, significantly reducing computational cost while maintaining the accuracy of the target distribution.
We evaluate this approach in Section~\ref{multieval} and discuss its effectiveness and limitations.

\section{Experiments}
\label{sec:experiments}
We evaluate ESS-Flow on generating materials with target properties, designing protein backbone structures, and drug-like molecules.
For material generation, we consider differentiable surrogate potentials to compare against state-of-the-art gradient-based methods including D-Flow \citep{ben2024d}, PnP-Flow \citep{martin2025pnp}, and DAPS \citep{zhang2025improving}.
For designing protein structures and molecules, the rewards are computed with non-differentiable external programs, which rules out gradient-based methods, and we compare against unconditional sampling from the prior.

\subsection{Material generation}
\label{sec:material}
We use FlowMM \citep{miller2024flowmm}, a flow-based generative model trained with Riemannian flow matching on the MP-20 dataset \citep{jain2013commentary}, as the prior over materials.
A crystalline material $c$ with unit lattice consisting of $n$ atoms is represented by the tuple $(\bm{a}, \bm{f}, \bm{l}, \bm{\beta})$, where $\bm{a} \in \{-1, 1\}^{n \times 7}$ are the binarized atomic numbers, $\bm{f} \in [0, 1]^{n \times 3}$ are the fractional coordinates of each atom, $\bm{l} \in \mathbb{R}_{+}^{3}$ are the lattice cell lengths, and $\bm{\beta} \in [60^\circ, 120^\circ]^{3}$ are the lattice angles.

While $\bm{f}, \bm{l}, \bm{\beta}$ are continuous variables, the 7-bit binarized atomic numbers $\bm{a}$ are discrete. 
FlowMM handles this by embedding atomic types in a continuous space and produces a soft encoding $\Tilde{\bm{a}} \in \mathbb{R}^{n \times 7}$, which is then rounded to obtain the final discrete $\bm{a}$. This rounding step makes the generative process non-differentiable with respect to the source. We also convert the uniform and log-normal source distributions of $\bm{f}, \bm{l}, \bm{\beta}$ into a standard Gaussian via a change of variables.

To demonstrate ESS-Flow's effectiveness with non-differentiable potentials, we first target materials with a specific space-group symmetry.
The potential function for this task is a binary indicator computed using an external program \citep{togo2024spglib}, making gradient-based methods inapplicable.
To enable comparison with gradient-based baselines, we additionally consider materials with target property values, where the properties are predicted using differentiable ALIGNN \citep{choudhary2021atomistic} models trained on the JARVIS-DFT \citep{choudhary2020joint} dataset rather than simulation-based procedures.
As target properties, we consider bulk modulus, shear modulus, and band gap, choosing target values above the 99th percentile of the prior property distribution.
Additionally, to evaluate the reward-tilted scenario, we guide generation toward stable materials (low predicted energy above hull) without targeting a specific value.
Since FlowMM requires specifying the number of atoms per unit lattice, we choose values based on the distribution of atoms for materials with large property values.
The target property values, number of atomic sites per lattice, and potential function details are summarized in Table \ref{tab:material}.

\begin{table}[t!]
\centering
\caption{Target properties and values for the material generation experiments.}
\label{tab:material}
\begin{small}
\begin{tabular}{@{}llcll@{}}
\toprule
Property $P_c$ & Target $y$ & Atoms $n$ & Potential function $g(c)$ & Scale $\sigma_y$ \\ \midrule
Space group & P6\textsubscript{3}/mmc & 8 & $\mathbf{1}[P_c = y]$ & -- \\
Bulk modulus & 300 GPa & 4 & $\exp(- (P_c - y)^2/ 2\sigma_y^2)$ & 10 GPa \\
Shear modulus & 200 GPa & 4 & $\exp(- (P_c - y)^2/ 2\sigma_y^2)$ & 10 GPa \\
Band gap & 10 eV & 12 & $\exp(- (P_c - y)^2/ 2\sigma_y^2)$ & 0.1 eV \\
Energy above hull & -- & 8 & $\exp(- P_c / \sigma_y)$ & 0.01 eV \\ \bottomrule
\end{tabular}
\end{small}
\end{table}

We apply dimension-wise learning rates for all methods, as gradient magnitudes vary across $(\bm{a}, \bm{f}, \bm{l}, \bm{\beta})$.
Gradient-based methods require approximations to handle the discrete atomic types.
ALIGNN expects integer atomic numbers $k_i$ for each atom, which it uses to create atom embeddings $\bm{h}_i$ by selecting the $k_i$-th row of its embedding matrix $\mathbf{E}$.  
For D-Flow and PnP-Flow, we maintain differentiability by using the continuous soft encoding $\Tilde{\bm{a}}_i$ to derive continuous values $\Tilde{k}_i$, and create soft atom embeddings $\bm{h}_i$:
\begin{equation}
    z_{ij} = (j - \Tilde{k}_i)^2, \quad \bm{v}_i = \mathrm{softmax}(-\bm{z}_i/\tau), \quad \bm{h}_i = \mathbf{E}^\mathsf{T}\bm{v}_i
\label{eq:atomapprox}
\end{equation}
where we set $\tau=0.1$.
For DAPS, we avoid this approximation by applying Langevin Monte Carlo only to the continuous variables $(\bm{f}, \bm{l}, \bm{\beta})$ and sampling $\bm{a}$ with Metropolis--Hastings using proposals that flip one bit of the binarized atomic numbers.
The noising and denoising steps are adapted from the original diffusion formulation to match FlowMM.
ESS-Flow requires no such approximations and handles discrete atomic types naturally, as it only makes point-wise evaluations of the generative model and potential function.
We provide a detailed analysis on the runtime costs of the different methods in Appendix~\ref{sec:cost}.

\begin{figure*}[t]
  \centering
  \begin{subfigure}[htbp]{0.9\textwidth}
    \centering
    \includegraphics[trim={0 2mm 0 0}, clip, width=\textwidth]{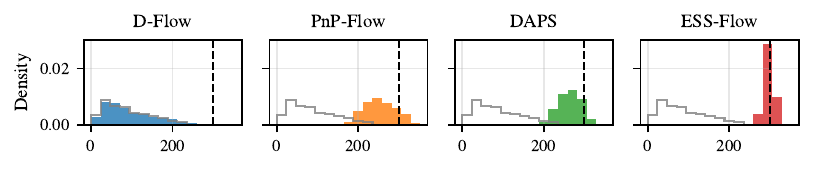}
    \caption{Bulk modulus: 300 GPa}
  \end{subfigure}
  \vfill
  \begin{subfigure}[htbp]{0.9\textwidth}
      \centering
      \includegraphics[trim={0 2mm 0 0}, clip, width=\textwidth]{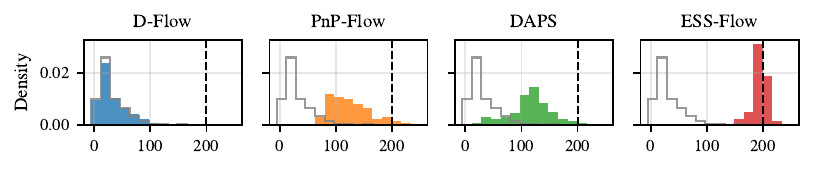}
      \caption{Shear modulus: 200 GPa}
  \end{subfigure}
  \vfill
  \begin{subfigure}[htbp]{0.9\textwidth}
      \centering
      \includegraphics[trim={0 2mm 0 0}, clip, width=\textwidth]{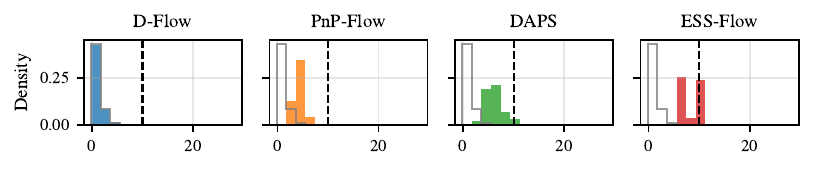}
      \caption{Band gap: 10 eV}
  \end{subfigure}
  \vfill
    \begin{subfigure}[htbp]{0.9\textwidth}
      \centering
      \includegraphics[trim={0 2mm 0 0}, clip, width=\textwidth]{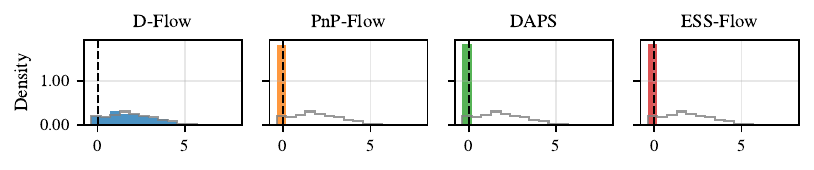}
      \caption{Energy above hull \textless 0 eV}
  \end{subfigure}
  \caption{Sample property distributions for different methods. Property values are along the x-axis, the prior property distribution is shown in gray, and target values are indicated by dotted lines. ESS-Flow samples concentrate near target values across all tasks.}
  \label{fig:prophist}
\end{figure*}

\paragraph{Results.}
ESS-Flow successfully generates 92.3\% of samples with the target P6\textsubscript{3}/mmc space group, compared to only 2.5\% from unconditional sampling.
For the property-targeting tasks, we measure performance using absolute errors between sample properties and targets, and for stability, we report the predicted energy above hull, as shown in Table~\ref{tab:res1}.
As anticipated from the toy example in Figure~\ref{fig:moons}, D-Flow fails to explore atomic compositions far from initialization even with the continuous approximation for $\bm{a}$, with performance nearly identical to unconditional sampling from the prior.
PnP-Flow, which optimizes in data space with noising-denoising steps, is less constrained by disconnected modes.
By avoiding gradient-based exploration of $\bm{a}$, DAPS and ESS-Flow perform considerably better. 
ESS-Flow achieves the lowest errors across all tasks, outperforming the baselines significantly.
The sample property distributions illustrated in Figure~\ref{fig:prophist} show that ESS-Flow better recovers the sharp target distributions.

\begin{table}[t]
\centering
\caption{Mean and standard deviation of absolute errors between the sample properties and targets, and of energy above hull values.}
\label{tab:res1}
\begin{small}
\begin{tabular}{@{}lcccc@{}}
\toprule
Method & Bulk modulus & Shear modulus & Band gap & Energy above hull \\ \midrule
D-Flow & 205.88 $\pm$ 60.79 & 165.93 $\pm$ 27.72 & 9.24 $\pm$ 1.18 & 1.92 $\pm$ 1.38 \\
PnP-Flow & 49.93 $\pm$ 32.33 & 75.48 $\pm$ 33.45 & 5.63 $\pm$ 1.03 & $-$0.02 $\pm$ 0.06 \\
DAPS & 39.14 $\pm$ 26.47 & 84.33 $\pm$ 37.10 & 3.90 $\pm$ 1.67 & $-$0.06 $\pm$ 0.05 \\
ESS-Flow & \textbf{8.99 $\pm$ 6.69} & \textbf{10.53 $\pm$ 9.21} & \textbf{1.85 $\pm$ 1.66} & \textbf{$-$0.19 $\pm$ 0.07} \\ \bottomrule
\end{tabular}
\end{small}
\end{table}

Beyond accuracy in matching target properties, we assess the quality of generated samples by computing the S.U.N.T. rate (stability, uniqueness, novelty, and threshold), adapted from prior works developing generative models for materials such as FlowMM and MatterGen \citep{zeni2025generative}. Briefly, the S.U.N.T. rate measures the following four properties: Stability (S.) as the proportion of structures that are thermodynamically stable after relaxation, Uniqueness (U.) as the proportion of generated structures that are distinct from each other, Novelty (N.) as the proportion that do not appear in the training set, and Threshold (T.) as the proportion meeting the desired target constraint.
The combined S.U.N.T. rate is the proportion of generated samples that satisfy all four criteria simultaneously. 
While the S.U.N. rate is an established quality metric for unconditional material generation, we introduce the threshold rate (T.) to adapt it to the controlled generation setting. 
A detailed description of each metric and the full relaxation pipeline can be found in Appendix~\ref{sec:matgen}.
The S.U.N.T. rates computed over 1000 generated samples are reported in Table~\ref{tab:quality1}.

\begin{table}[t]
\centering
\caption{S.U.N.T. rates (\%) for the generated samples across different material generation tasks.}
\label{tab:quality1}
\begin{small}
\begin{tabular}{@{}lccccc@{}}
\toprule
Method & Bulk modulus & Shear modulus & Band gap & Energy above hull & Space group \\ \midrule
Unconditional & 0.0 & 0.0 & 0.0 & 1.4 & 1.3 \\
D-Flow & 0.0 & 0.0 & 0.0 & 1.8 & -- \\
PnP-Flow & 3.8 & 0.9 & 0.1 & 34.5 & -- \\
DAPS & 9.4 & 0.5 & 0.0 & 17.6 & -- \\
ESS-Flow & \textbf{13.7} & \textbf{4.7} & \textbf{16.0} & \textbf{37.6} & \textbf{25.5} \\ \bottomrule
\end{tabular}
\end{small}
\end{table}

ESS-Flow successfully targets the desired properties and achieves the highest S.U.N.T. rates across all tasks.
For comparison, the S.U.N. rate for unconditional generation with FlowMM is nearly 26\%.
Despite the challenging controlled generation problem, where we target extreme values for the desired properties, the resulting S.U.N.T. rates indicate that ESS-Flow generates materials that are viable candidates for further exploration.

\subsubsection{Evaluation of multi-fidelity sampling} 
\label{multieval}
We perform a preliminary evaluation of the multi-fidelity sampling approach proposed in Section~\ref{sec:multifidelity}.
ESS-Flow samples are obtained by evaluating the potential with a coarse discretization $\Delta=1/50$, then reweighted using importance weights computed with a much finer discretization $\delta=1/1000$, as described in equation~\eqref{eq:importanceweight}. 
The weighted samples have reasonable effective sample sizes of 65.3\% and 33.9\% for the bulk modulus and shear modulus tasks, respectively.
However, for the band gap and stability tasks, which have sharper, concentrated target distributions, the effective sample sizes drop to 0.1\% and 1.0\%, respectively. 
This is a limitation of the simple importance reweighting approach, 
where the weights $w \propto g(T^{\delta}(z))/g(T^{\Delta}(z))$ can become extremely large when the coarse discretization yields samples with lower potential values.
Further analysis comparing ESS-Flow samples and the importance weighted samples is provided in Appendix \ref{comparemulti}.

\subsection{Protein backbone design}

We consider the 60M-parameter variant of Proteina \citep{geffner2025proteina}, a flow-based generative model, as the prior over protein backbones with 100 residues.
We target structures with higher designability, which is measured as follows \citep{pmlr-v202-yim23a, geffner2025proteina}.
For a generated backbone, $N$ candidate amino acid sequences are predicted with the inverse folding model ProteinMPNN \citep{dauparas2022robust} at temperature 0.1, and each sequence is refolded with ESMFold \citep{lin2023evolutionary}.
The self-consistency RMSD (scRMSD) is defined as the minimum RMSD between the refolded structures and the generated backbone after alignment.
The backbone is considered designable if its scRMSD is below 2\AA, and designability is the percentage of designable backbones among the generated samples.

We sample the tilted prior with the non-differentiable potential $g(x) = \exp(-w \cdot \text{scRMSD}(x))$, computed with $N=1$, for different reward strengths $w>0$.
All samples are generated with the probability flow ODE using 50 steps, and ESS-Flow is run with 16 parallel MCMC chains for 264 steps, discarding the first 200 steps as burn-in.
Over the resulting 1024 samples, we report the mean scRMSD and designability, evaluated with $N=8$.
We also measure diversity as the average pairwise TM-score \citep{geffner2025proteina} among the designable samples, where lower values indicate higher diversity.

\begin{table}[t]
\centering
\caption{Designability and diversity of generated protein backbone structures.}
\label{tab:proteina}
\begin{small}
\begin{tabular}{@{}lccc@{}}
\toprule
Method & scRMSD (\AA) $\downarrow$ & Designability (\%) $\uparrow$ & TM-score $\downarrow$ \\ \midrule
Unconditional & 6.74 & 9.6 & 0.361 \\
ESS-Flow ($w=0.5$) & 2.49 & 53.1 & 0.368 \\
ESS-Flow ($w=1.0$) & 1.92 & 74.0 & 0.367 \\
ESS-Flow ($w=2.0$) & 1.53 & 86.4 & 0.380 \\ \bottomrule
\end{tabular}
\end{small}
\end{table}

\paragraph{Results.}
As shown in Table \ref{tab:proteina}, ESS-Flow lowers the mean scRMSD and substantially improves designability over unconditional sampling, with larger gains at higher reward strengths.
Sample diversity is largely preserved, with a slight increase in the pairwise TM-score at the highest reward strength as the tilted prior becomes more concentrated.

\subsection{Molecule generation}
We use GeoLDM \citep{pmlr-v202-xu23n}, a latent diffusion model trained on the GEOM-Drugs dataset \citep{axelrod2022geom}, as the prior over 28-atom 3D molecules.
With the DDIM sampler using 50 steps, the composition of the latent transport and the decoder defines a deterministic map from the Gaussian source to molecules, making GeoLDM suitable for ESS-Flow.
We target synthesizable, drug-like molecules using the quantitative estimate of drug-likeness, $\text{QED} \in [0,1]$, where higher values indicate higher drug-likeness, and the synthetic accessibility score, $\text{SA} \in [1,10]$, where lower values indicate ease of synthesis, both computed with the RDKit library.
Unconditional samples from GeoLDM often consist of disconnected fragments, and those that remain connected have low QED and high SA scores.
For a reward strength $w>0$, we sample the tilted prior with the non-differentiable potential $g(x) = \exp(-w)$ if the molecule $x$ is disconnected, and $g(x) = \exp(w \cdot (\text{QED}(x) + \text{SA}_\text{norm}(x))/2)$ otherwise, where $\text{SA}_\text{norm} = (10-\text{SA})/9 \in [0,1]$.

Following \citet{lee2025genmol}, we report the percentage of connected molecules (Conn.), and filter out disconnected ones before computing diversity (Div.), the percentage of unique molecules (Uniq.), and the QED and SA scores over 1024 generated samples.
Uniqueness is the percentage of distinct canonical SMILES, and diversity is the average pairwise Tanimoto distance between the Morgan fingerprints of all connected molecules.
We report quality (Qual.) as the percentage of samples that are connected and unique, with $\text{QED} \geq 0.6$ and $\text{SA} \leq 4.0$. 
We compare against unconditional sampling with both the original DDPM sampler with 1000 steps and the deterministic DDIM sampler.
ESS-Flow is run with the same configuration as in the protein backbone design experiment.

\begin{table}[t]
\centering
\caption{Metrics for the generated molecules. Conn., Uniq., and Qual. are in percentage, and QED and SA scores are reported with mean and standard deviation.}
\label{tab:geoldm}
\begin{small}
\begin{tabular}{@{}lcccccc@{}}
\toprule
Method & Conn.\ $\uparrow$ & Div.\ $\uparrow$ & Uniq.\ $\uparrow$ & QED $\uparrow$ & SA $\downarrow$ & Qual.\ $\uparrow$ \\ \midrule
Unconditional (DDPM) & 70.0 & 0.843 & 100.0 & 0.413 $\pm$ 0.183 & 6.954 $\pm$ 0.934 & 0.0 \\
Unconditional (DDIM) & 44.3 & 0.848 & 100.0 & 0.351 $\pm$ 0.177 & 6.922 $\pm$ 0.945 & 0.1 \\
ESS-Flow ($w=5$) & 99.8 & 0.849 & 91.6 & 0.435 $\pm$ 0.189 & 6.689 $\pm$ 1.116 & 0.3 \\
ESS-Flow ($w=10$) & 99.1 & 0.848 & 80.4 & 0.520 $\pm$ 0.190 & 6.377 $\pm$ 1.336 & 0.4 \\
ESS-Flow ($w=20$) & 97.9 & 0.826 & 39.7 & 0.635 $\pm$ 0.159 & 4.760 $\pm$ 2.252 & 0.9 \\ \bottomrule
\end{tabular}
\end{small}
\end{table}

\paragraph{Results.}
ESS-Flow almost exclusively produces connected molecules as shown in Table \ref{tab:geoldm}, compared to unconditional generation with both DDPM and DDIM samplers.
The QED and SA scores steadily improve with increasing reward strength, and at $w=20$, the fraction of unique molecules drops as the tilted prior becomes more concentrated.
However, the overall quality, which accounts for uniqueness and the QED and SA thresholds, continues to increase with reward strength, indicating that ESS-Flow reliably steers generation towards the desired properties.

\section{Limitations and broader impacts}
\label{sec:limitations}
As a method for sampling the tilted prior, the performance of ESS-Flow is limited by how well the generative model approximates the true data distribution under moderate ODE discretization. Recent developments in flow maps \cite{boffibuild} that maintain sample quality with few step evaluations will further enable the method.
Additionally, ESS-Flow is more expensive in terms of the number of function evaluations than the compared baselines, which should be expected of a zeroth-order sampling method designed to solve a broader class of problems where gradient-based methods fail. 
However, we have not optimized the method for computational efficiency in this work. 
ESS admits tunable variants that reduce the number of shrinkage steps per iteration, by restricting the initial angular bracket to a length smaller than $2\pi$, or by biasing proposals toward the current state instead of sampling uniformly along the ellipse.
Such problem-specific tuning moves the method closer to pCN, at the expense of the tuning-free nature of ESS-Flow.
Lastly, warm starting the MCMC chains and leveraging parallel chains to speed up mixing \cite{nishihara2014parallel} would further reduce the computational cost.

Controlled generation methods can accelerate the discovery of therapeutic drugs and novel materials by improving the screening process.
However, they carry dual-use risks, as they could be repurposed to design harmful substances \cite{urbina2022dual}.
While generative models have wide-ranging societal impacts, ESS-Flow does not introduce new risks beyond those of the pretrained models it builds upon.

\section{Conclusion}
We present ESS-Flow, a training-free and gradient-free sampling method for controlled generation with pretrained flow-based generative models.
The method is particularly valuable for problems where gradients are unavailable, such as when the generative process involves quantization or when evaluating the potential requires non-differentiable simulations, settings that are common in scientific applications.
We propose and evaluate a multi-fidelity extension using importance reweighting as a proof of concept.
Future work could explore more sophisticated multi-fidelity approaches, such as tempering-based methods, to better balance computational efficiency with sampling accuracy.

\begin{ack}
This work was financially supported by the Wallenberg AI, Autonomous Systems and Software Program (WASP) funded by the Knut and Alice Wallenberg Foundation, the Swedish Research Council (project no: 2024-05011), and the Excellence Center at Linköping--Lund in Information Technology (ELLIIT). Computational resources were provided on the Berzelius system funded by the Knut and Alice Wallenberg foundation and operated by NAISS.
\end{ack}

\bibliographystyle{plainnat-mod}
\bibliography{neurips_2026}

\section*{NeurIPS Paper Checklist}

\begin{enumerate}

\item {\bf Claims}
    \item[] Question: Do the main claims made in the abstract and introduction accurately reflect the paper's contributions and scope?
    \item[] Answer: \answerYes{} 
    \item[] Justification: We claim our main contribution as introducing a training-free method for controlled generation in flow-based models that does not require differentiability of the generative or observation processes. We present the method in detail in Section~\ref{sec:method} and provide experimental validation in Section~\ref{sec:experiments}.

\item {\bf Limitations}
    \item[] Question: Does the paper discuss the limitations of the work performed by the authors?
    \item[] Answer: \answerYes{} 
    \item[] Justification: We discuss the limitations of our work in Section \ref{sec:limitations}.

\item {\bf Theory assumptions and proofs}
    \item[] Question: For each theoretical result, does the paper provide the full set of assumptions and a complete (and correct) proof?
    \item[] Answer: \answerNA{} 
    \item[] Justification: While we restate the result of \citet{Natarovskii2021} adapted to our setting for completeness, we do not claim any new theoretical contribution. The consistency and convergence properties of our method rely on existing theoretical results for elliptical slice sampling.

\item {\bf Experimental result reproducibility}
    \item[] Question: Does the paper fully disclose all the information needed to reproduce the main experimental results of the paper to the extent that it affects the main claims and/or conclusions of the paper (regardless of whether the code and data are provided or not)?
    \item[] Answer: \answerYes{}{} 
    \item[] Justification: The experiments and evaluation metrics are described in Section~\ref{sec:experiments}, with implementation details for all methods provided in the Appendix.

\item {\bf Open access to data and code}
    \item[] Question: Does the paper provide open access to the data and code, with sufficient instructions to faithfully reproduce the main experimental results, as described in supplemental material?
    \item[] Answer: \answerNo{} 
    \item[] Justification: The code is not provided at the time of submission, but will be made publicly available upon publication.

\item {\bf Experimental setting/details}
    \item[] Question: Does the paper specify all the training and test details (e.g., data splits, hyperparameters, how they were chosen, type of optimizer) necessary to understand the results?
    \item[] Answer: \answerYes{} 
    \item[] Justification: A general description of the experiments is provided in Section~\ref{sec:experiments}, with additional hyperparameter details included in the Appendix.

\item {\bf Experiment statistical significance}
    \item[] Question: Does the paper report error bars suitably and correctly defined or other appropriate information about the statistical significance of the experiments?
    \item[] Answer: \answerYes{} 
    \item[] Justification: We report the mean and standard deviation of the metrics across generated samples in the results.

\item {\bf Experiments compute resources}
    \item[] Question: For each experiment, does the paper provide sufficient information on the computer resources (type of compute workers, memory, time of execution) needed to reproduce the experiments?
    \item[] Answer: \answerYes{} 
    \item[] Justification: We compare and discuss the computational costs and peak GPU memory usage for all the methods in the Appendix.
    
\item {\bf Code of ethics}
    \item[] Question: Does the research conducted in the paper conform, in every respect, with the NeurIPS Code of Ethics \url{https://neurips.cc/public/EthicsGuidelines}?
    \item[] Answer: \answerYes{} 
    \item[] Justification: We have reviewed the guidelines and confirm that our work adheres to the NeurIPS Code of Ethics.

\item {\bf Broader impacts}
    \item[] Question: Does the paper discuss both potential positive societal impacts and negative societal impacts of the work performed?
    \item[] Answer: \answerYes{} 
    \item[] Justification: We discuss the broader impacts of the work in Section~\ref{sec:limitations}.
    
\item {\bf Safeguards}
    \item[] Question: Does the paper describe safeguards that have been put in place for responsible release of data or models that have a high risk for misuse (e.g., pre-trained language models, image generators, or scraped datasets)?
    \item[] Answer: \answerNA{} 
    \item[] Justification: We do not release any new model or dataset in this work.

\item {\bf Licenses for existing assets}
    \item[] Question: Are the creators or original owners of assets (e.g., code, data, models), used in the paper, properly credited and are the license and terms of use explicitly mentioned and properly respected?
    \item[] Answer: \answerYes{} 
    \item[] Justification: All existing datasets, models, and code used in this work are appropriately cited, and are used in accordance with their respective licenses and terms of use.

\item {\bf New assets}
    \item[] Question: Are new assets introduced in the paper well documented and is the documentation provided alongside the assets?
    \item[] Answer: \answerNA{}{} 
    \item[] Justification: No new datasets or models are released with this work. The implementation code will be made public upon publication.

\item {\bf Crowdsourcing and research with human subjects}
    \item[] Question: For crowdsourcing experiments and research with human subjects, does the paper include the full text of instructions given to participants and screenshots, if applicable, as well as details about compensation (if any)? 
    \item[] Answer: \answerNA{} 
    \item[] Justification: This work does not involve crowdsourcing experiments or research with human subjects.

\item {\bf Institutional review board (IRB) approvals or equivalent for research with human subjects}
    \item[] Question: Does the paper describe potential risks incurred by study participants, whether such risks were disclosed to the subjects, and whether Institutional Review Board (IRB) approvals (or an equivalent approval/review based on the requirements of your country or institution) were obtained?
    \item[] Answer: \answerNA{} 
    \item[] Justification: The paper does not involve crowdsourcing or research with human subjects.

\item {\bf Declaration of LLM usage}
    \item[] Question: Does the paper describe the usage of LLMs if it is an important, original, or non-standard component of the core methods in this research? Note that if the LLM is used only for writing, editing, or formatting purposes and does \emph{not} impact the core methodology, scientific rigor, or originality of the research, declaration is not required.
    \item[] Answer: \answerNA{} 
    \item[] Justification: The core method development does not involve LLMs.

\end{enumerate}

\newpage
\appendix

\section{Gaussian mixture model}
\label{gmm}
We numerically evaluate how ESS-Flow scales with dimension on simulated data.
The prior distribution is a Gaussian mixture model with 25 components, with means $\mu_{i,j} = (i, j, \ldots, i, j) \in \mathbb{R}^{d_x}$ for $(i, j) \in \{-2, -1, 0, 1, 2\}^2$ and covariances $(1/8)^2 I_{d_x}$.
The unnormalized weights of the mixture components are independently drawn from a $\chi^2$ distribution.
We consider a linear-Gaussian observation model $y = Ax + \sigma_y \epsilon$ with $\epsilon \sim \mathcal{N}(0, I_{d_y})$, so the posterior is also a Gaussian mixture that can be computed exactly.
The setup is similar to that of \citet{cardoso2024monte}, except we use rescaled means and covariances while maintaining the relative separation between mixture components.

To avoid model approximation errors from learning a transport map, we use the analytical expression derived in \citet{albergo2023stochastic} for the linear interpolant $x_t = (1-t) x_0 + tx_1$ from a standard normal source to the Gaussian mixture.
For ESS-Flow, we run 1000 parallel MCMC chains for 501 steps each and collect the last sample from each chain.
The ODE is numerically integrated using the midpoint method with a step size of 0.05.
We report the sliced Wasserstein distance (SWD) between ESS-Flow samples and exact posterior samples in Table \ref{tab:gmm}.
The average number of likelihood evaluations per MCMC step, denoted by $n_{\text{ellipse}}$, measures sampling efficiency within ESS.

\begin{table}[t]
\centering
\caption{Sliced Wasserstein distance (mean $\pm$ standard error) and average number of likelihood evaluations per MCMC step (mean $\pm$ standard deviation) across 10 random seeds.}
\label{tab:gmm}
\begin{small}
\begin{tabular}{@{}ccccccc@{}}
\toprule
 & \multicolumn{3}{c}{SWD} & \multicolumn{3}{c}{$n_{\text{ellipse}}$} \\ \cmidrule(lr){2-4} \cmidrule(l){5-7}
$d_y$ & $d_x = 8$ & $d_x = 80$ & $d_x = 800$ & $d_x = 8$ & $d_x = 80$ & $d_x = 800$ \\ \midrule
1 & 0.037 $\pm$ 0.010 & 0.044 $\pm$ 0.019 & 0.064 $\pm$ 0.025 & 2.8 $\pm$ 1.3 & 2.8 $\pm$ 1.1 & 2.7 $\pm$ 0.5 \\
2 & 0.034 $\pm$ 0.014 & 0.034 $\pm$ 0.009 & 0.037 $\pm$ 0.016 & 3.2 $\pm$ 0.7 & 2.8 $\pm$ 1.2 & 3.2 $\pm$ 1.6 \\
4 & 0.028 $\pm$ 0.008 & 0.031 $\pm$ 0.014 & 0.038 $\pm$ 0.024 & 3.5 $\pm$ 1.2 & 3.0 $\pm$ 1.4 & 4.1 $\pm$ 2.4 \\ \bottomrule
\end{tabular}
\end{small}
\end{table}

\begin{figure}[t]
\begin{center}
\includegraphics[width=0.7\linewidth]{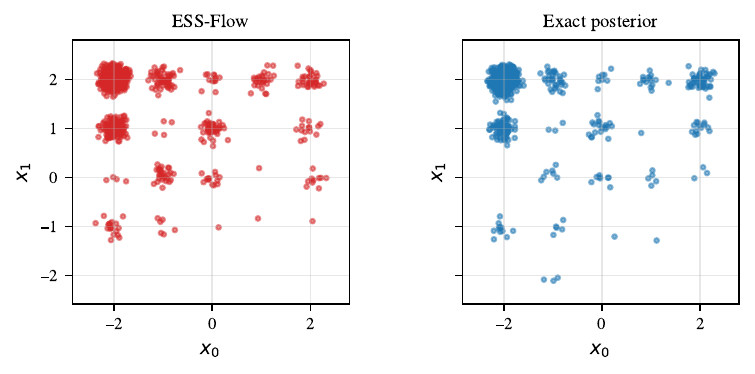}
\end{center}
\caption{First two dimensions of the posterior approximated with ESS-Flow compared to the exact posterior for the GMM example with $d_x=800$ and $d_y=1$.}
\label{fig:gmm}
\end{figure}

As dimension increases, ESS-Flow's sampling efficiency remains stable and the SWD remains low, indicating good approximation. 
For qualitative evaluation, we visualize the first two dimensions of the posterior for one random seed with $d_x = 800$ and $d_y=1$ in Figure \ref{fig:gmm}.

\section{Additional details for material generation}
The hyperparameters for all methods used in the material generation experiments are provided in Table \ref{tab:hyp1}. 

\begin{table}[t]
\centering
\caption{Hyperparameters for the material generation tasks. Gradient-based methods use dimension-specific learning rates for $(\bm{a}, \bm{f}, \bm{l}, \bm{\beta})$.}
\label{tab:hyp1}
\begin{small}
\begin{tabular}{@{}lcccc@{}}
\toprule
Method / Parameter & Bulk modulus & Shear modulus & Band gap & Energy above hull \\ \midrule
\textbf{D-Flow} &  &  &  &  \\
ODE integration steps $N$ & 50 & 50 & 50 & 50 \\
Optimizer & SGD & SGD & SGD & SGD \\
Optimization steps $N_{opt}$ & 100 & 100 & 100 & 100 \\
Learning rate $(\bm{a}, \bm{l}, \bm{\beta})$ & $10^{-6}$ & $10^{-5}$ & $10^{-4}$ & $10^{-4}$ \\
Learning rate $(\bm{f})$ & $10^{-8}$ & $10^{-7}$ & $10^{-6}$ & $10^{-6}$ \\ \midrule
\textbf{PnP-Flow} &  &  &  &  \\
Optimizer & SGD & SGD & SGD & SGD \\
Optimization steps $N_{opt}$ & 100 & 100 & 100 & 100 \\
Learning rate decay & linear & linear & linear & linear \\
Min. learning rate & 0 & 0 & 0 & 0 \\
Max. learning rate $(\bm{a})$ & $10^{-4}$ & $10^{-3}$ & $10^{-1}$ & $10^{-2}$ \\
Max. learning rate $(\bm{f})$ & $10^{-6}$ & $10^{-5}$ & $10^{-3}$ & $10^{-4}$ \\
Max. learning rate $(\bm{l}, \bm{\beta})$ & $10^{-5}$ & $10^{-4}$ & $10^{-2}$ & $10^{-3}$ \\
ODE integration steps $N$ & $100 - n$ & $100 - n$ & $100 - n$ & $100 - n$ \\ \midrule
\textbf{DAPS} &  &  &  &  \\
ODE integration steps $N$ & 5 & 5 & 5 & 5 \\
Optimization steps $N_{opt}$ & 100 & 100 & 100 & 100 \\
Sampling steps $N_s$ & 20 & 20 & 20 & 20 \\
Learning rate decay & linear & linear & linear & linear \\
Min. learning rate & $10^{-2} \alpha$ & $10^{-2} \alpha$ & $10^{-2} \alpha$ & $10^{-2} \alpha$ \\
Max. learning rate $(\alpha_{\bm{f}})$ & $10^{-3}$ & $10^{-3}$ & $10^{-5}$ & $10^{-4}$ \\
Max. learning rate $(\alpha_{\bm{l}}, \alpha_{\bm{\beta}})$ & $10^{-2}$ & $10^{-2}$ & $10^{-4}$ & $10^{-3}$ \\
$p(x_0|x_t)$ scale $(r_{\bm{a}})$ & $3 (1 - t)$ & $3 (1 - t)$ & $3 (1 - t)$ & $3 (1 - t)$ \\
$p(x_0|x_t)$ scale $(r_{\bm{f}}, r_{\bm{l}}, r_{\bm{\beta}})$ & $1 - t$ & $1 - t$ & $1 - t$ & $1 - t$ \\ \midrule
\textbf{ESS-Flow} &  &  &  &  \\
ODE integration steps $N$ & 50 & 50 & 50 & 50 \\
MCMC steps & 201 & 201 & 501 & 201 \\
Burn-in & 200 & 200 & 500 & 200 \\ \bottomrule
\end{tabular}
\end{small}
\end{table}

\subsection{Quality metrics}
\label{sec:matgen}
We describe how the S.U.N.T. rate is computed for a set of $M$ generated materials $\{c_1, \dots, c_M\}$.

\paragraph{Validity and relaxation.}
A generated material $c_m$ is valid if its lattice volume is greater than 0.1 \AA$^3$, its atomic numbers are within the valid range, and its charge is balanced.
\begin{equation}
    v_m = \mathbf{1}[\text{vol}(c_m) > 0.1] \cdot \mathbf{1}[a_m \in \mathcal{A}_{\text{valid}}] \cdot \mathbf{1}[\text{charge}(c_m) = 0],
\end{equation}
where $\mathcal{A}_{\text{valid}}$ denotes the set of valid atomic numbers. We discard invalid materials and relax the remaining structures using CHGNet \citep{deng2023chgnet}, following FlowMM, without further relaxations using density functional theory. Let $\tilde{c}_m$ denote the relaxed structure.

\paragraph{Stability.}
The stability indicator for a relaxed structure $\tilde{c}_m$ is given by
\begin{equation}
    s_m = \mathbf{1}[E_\text{hull}(\tilde{c}_m) < 0.1 \text{ eV/atom}],
\end{equation}
where $E_\text{hull}$ is the energy above hull predicted by CHGNet.
Note that this differs from the ALIGNN predictor used to guide generation of metastable materials, and can thus produce different results. The stability rate (S.) is $\sum_m{v_m s_m}/M$.

\paragraph{Uniqueness and novelty.}
Two relaxed structures $\tilde{c}_m$ and $\tilde{c}_{m'}$ are considered identical if they are matched by pymatgen's \texttt{StructureMatcher} \citep{ong2013python} with default settings.
The uniqueness and novelty indicators are given by
\begin{align}
    u_m &= \mathbf{1}[\tilde{c}_m \text{ is not matched by any } \tilde{c}_{m'}, m' \neq m], \\
    n_m &= \mathbf{1}[\tilde{c}_m \text{ is not matched by any structure in the training set}].
\end{align}
The uniqueness (U.) and novelty (N.) rates are $\sum_m v_mu_m/M$ and $\sum_m v_mn_m/M$, respectively.

\paragraph{Threshold.}
The threshold rate measures the proportion of generated materials meeting the desired target constraint for the specific task. The threshold indicator $t_m$ is task specific: for bulk modulus, shear modulus, and band gap, it requires the predicted property to be within 5\% of the target value; for energy above hull, it requires the predicted value to be less than 0 eV; and for space group, it requires an exact match with the target space group P6\textsubscript{3}/mmc. The threshold rate (T.) is $\sum_m v_m t_m/M$.

\paragraph{Combined S.U.N.T. rate.}
The S.U.N.T. rate is the proportion of generated materials satisfying all four criteria.
\begin{equation}
\text{S.U.N.T.} = \frac{1}{M}\sum_{m=1}^{M} v_m \cdot s_m \cdot u_m \cdot n_m \cdot t_m.
\end{equation}

A breakdown of these metrics for the generated samples is given in Table \ref{tab:quality}.

\begin{table}[t]
\centering
\caption{Quality metrics (\%) for generated samples across different material generation tasks.}
\label{tab:quality}
\begin{small}
\begin{tabular}{@{}llcccccc@{}}
\toprule
Task & Method & Valid & S. & U.N. & T. & S.U.N. & S.U.N.T. \\ \midrule
\multirow{2}{*}{Space group} & Unconditional & 73.0 & 26.5 & 72.4 & 2.3 & 26.0 & 1.3 \\
 & ESS-Flow & 87.1 & 50.0 & 54.6 & 81.9 & 26.7 & \textbf{25.5} \\ \midrule
\multirow{5}{*}{Bulk modulus} & Unconditional & 77.9 & 32.4 & 73.2 & 0.4 & 28.3 & 0.0 \\
 & D-Flow & 75.2 & 31.0 & 70.2 & 0.4 & 26.3 & 0.0 \\
 & PnP-Flow & 94.8 & 46.2 & 71.4 & 17.8 & 29.0 & 3.8 \\
 & DAPS & 95.4 & 57.6 & 80.8 & 19.8 & 47.0 & 9.4 \\
 & ESS-Flow & 98.3 & 56.4 & 46.1 & 79.6 & 19.3 & \textbf{13.7} \\ \midrule
\multirow{5}{*}{Shear modulus} & Unconditional & 76.6 & 31.4 & 71.9 & 0.0 & 27.3 & 0.0 \\
 & D-Flow & 78.6 & 32.2 & 73.2 & 0.0 & 27.5 & 0.0 \\
 & PnP-Flow & 92.7 & 43.2 & 68.3 & 5.4 & 25.7 & 0.9 \\
 & DAPS & 86.2 & 47.2 & 74.6 & 1.8 & 39.1 & 0.5 \\
 & ESS-Flow & 98.2 & 50.3 & 30.5 & 59.9 & 10.0 & \textbf{4.7} \\ \midrule
\multirow{5}{*}{Band gap} & Unconditional & 68.0 & 14.5 & 68.0 & 0.0 & 14.5 & 0.0 \\
 & D-Flow & 69.8 & 16.2 & 69.7 & 0.0 & 16.1 & 0.0 \\
 & PnP-Flow & 44.0 & 1.6 & 43.8 & 0.2 & 1.6 & 0.1 \\
 & DAPS & 7.1 & 0.4 & 7.1 & 0.0 & 0.4 & 0.0 \\
 & ESS-Flow & 48.8 & 16.9 & 48.0 & 23.2 & 16.5 & \textbf{16.0} \\ \midrule
\multirow{5}{*}{Energy above hull} & Unconditional & 69.9 & 25.1 & 69.0 & 2.3 & 24.2 & 1.4 \\
 & D-Flow & 69.7 & 25.2 & 68.8 & 2.7 & 24.3 & 1.8 \\
 & PnP-Flow & 79.8 & 52.6 & 73.9 & 57.4 & 47.2 & 34.5 \\
 & DAPS & 47.0 & 19.9 & 46.8 & 42.0 & 19.7 & 17.6 \\
 & ESS-Flow & 73.3 & 44.5 & 64.9 & 73.3 & 37.6 & \textbf{37.6} \\ \bottomrule
\end{tabular}
\end{small}
\end{table}

\subsection{Runtime cost analysis}
\label{sec:cost}
We report the number of function evaluations (NFE) of both the generative model $T_\theta$ and the potential function $g$ for all the methods in Tables \ref{tab:compute_cost_baselines} and \ref{tab:compute_cost_essflow}. For ESS-Flow, the NFE varies based on the number of evaluations made along the ellipse per MCMC step ($n_{\text{ellipse}}$), where we report the mean values for each task, averaged over 1000 independent MCMC chains. Unlike ESS-Flow, gradient-based methods additionally require backward passes through the generative model, the potential function, or both, further increasing the computational cost, which is not reflected in this metric.

\begin{table}[t]
\centering
\caption{Number of function evaluations per generated sample for the baseline methods.}
\label{tab:compute_cost_baselines}
\begin{small}
\begin{tabular}{@{}lccc@{}}
\toprule
Method & $\text{NFE}_{T_\theta}$ & $\text{NFE}_g$ & Gradients \\ \midrule
D-Flow & $5 \times 10^3$ & $1 \times 10^2$ & $T_\theta$ and $g$ \\
PnP-Flow & $5 \times 10^3$ & $1 \times 10^2$ & $g$ only \\
DAPS & $5 \times 10^2$ & $2 \times 10^3$ & $g$ only \\ \bottomrule
\end{tabular}
\end{small}
\end{table}

\begin{table}[t]
\centering
\caption{Number of function evaluations per generated sample for ESS-Flow on each task.}
\label{tab:compute_cost_essflow}
\begin{small}
\begin{tabular}{@{}lcccc@{}}
\toprule
Task & $n_{\text{ellipse}}$ & $\text{NFE}_{T_\theta}$ & $\text{NFE}_g$ & Gradients \\ \midrule
Bulk modulus & 6.7 & $7 \times 10^4$ & $1 \times 10^3$ & None \\
Shear modulus & 7.1 & $7 \times 10^4$ & $1 \times 10^3$ & None \\
Band gap & 17.4 & $4 \times 10^5$ & $9 \times 10^3$ & None \\
Energy above hull & 12.0 & $1 \times 10^5$ & $2 \times 10^3$ & None \\
Space group & 4.8 & $5 \times 10^4$ & $1 \times 10^3$ & None \\ \bottomrule
\end{tabular}
\end{small}
\end{table}

The NFE for ESS-Flow is an order of magnitude higher than the baseline methods, but this should be expected of a zeroth-order sampling method designed to solve a broader class of problems where gradient-based methods fail. We have also not optimized the method for computational efficiency in this work. In the material generation experiments, we run parallel MCMC chains and treat only the final step as the generated sample, discarding all preceding steps as burn-in. Instead, retaining multiple thinned samples from each chain would lower the NFE per sample.

The overall computational cost can be further reduced by running ESS-Flow in parallel with large batch sizes, enabled by its low memory requirements.
Table \ref{tab:memory} shows the peak GPU memory usage for the different methods with a batch size of 50. ESS-Flow requires up to two orders of magnitude less memory than D-Flow.

\begin{table}[t]
\centering
\caption{Peak GPU memory usage (GB) for the different methods on material generation tasks.}
\label{tab:memory}
\begin{small}
\begin{tabular}{@{}lccccc@{}}
\toprule
Method & Bulk modulus & Shear modulus & Band gap & Energy above hull & Space group \\ \midrule
D-Flow & 6.91 & 6.91 & 62.22 & 27.90 & -- \\
PnP-Flow & 0.86 & 0.86 & 2.28 & 1.57 & -- \\
DAPS & 0.86 & 0.86 & 2.28 & 1.57 & -- \\
ESS-Flow & 0.29 & 0.29 & 0.60 & 0.45 & 0.18 \\ \bottomrule
\end{tabular}
\end{small}
\end{table}

\subsection{Additional evaluation of multi-fidelity sampling}
\label{comparemulti}
\begin{figure}[t]
  \centering
  \begin{subfigure}[htbp]{0.45\textwidth}
    \centering
    \includegraphics[trim={0 2mm 0 0}, clip, width=\textwidth]{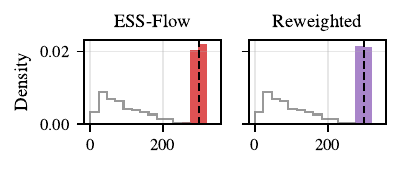}
    \caption{Bulk modulus: 300 GPa}
  \end{subfigure}
  \begin{subfigure}[htbp]{0.45\textwidth}
      \centering
      \includegraphics[trim={0 2mm 0 0}, clip, width=\textwidth]{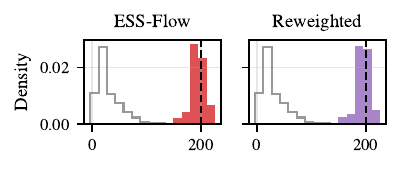}
      \caption{Shear modulus: 200 GPa}
  \end{subfigure}
  \begin{subfigure}[htbp]{0.45\textwidth}
      \centering
      \includegraphics[trim={0 2mm 0 0}, clip, width=\textwidth]{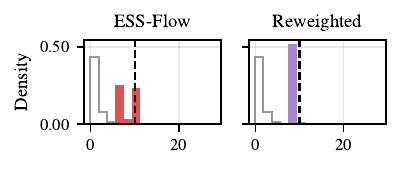}
      \caption{Band gap: 10 eV}
  \end{subfigure}
    \begin{subfigure}[htbp]{0.45\textwidth}
      \centering
      \includegraphics[trim={0 2mm 0 0}, clip, width=\textwidth]{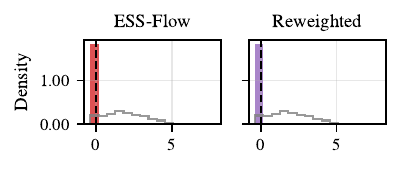}
      \caption{Energy above hull \textless 0 eV}
  \end{subfigure}
  \caption{Comparison of sample property distributions between standard ESS-Flow and importance reweighted ESS-Flow. Property values are on the x-axis, the prior property distribution is shown in gray, and target values are indicated by dotted lines.}
  \label{fig:multiprophist}
\end{figure}

We evaluate the distribution of material properties and quality of importance-reweighted samples compared to standard ESS-Flow samples.
We resample 1000 source samples according to their self-normalized importance weights and obtain the corresponding material structures by integrating the ODE with fine discretization $\delta = 1/1000$. 
The resulting property distributions are shown in Figure \ref{fig:multiprophist}, and the accuracy metrics are reported in Table \ref{tab:res1multi}.

\begin{table}[t]
\centering
\caption{Mean and standard deviation of absolute errors between the sample properties and targets, and of energy above hull values.}
\label{tab:res1multi}
\begin{small}
\begin{tabular}{@{}lcccc@{}}
\toprule
Method & Bulk modulus & Shear modulus & Band gap & Energy above hull \\ \midrule
ESS-Flow & 8.99 $\pm$ 6.69 & 10.53 $\pm$ 9.21 & 1.85 $\pm$ 1.66 & $-$0.19 $\pm$ 0.07 \\
Reweighted & 9.36 $\pm$ 6.70 & 10.00 $\pm$ 9.35 & 0.85 $\pm$ 0.47 & $-$0.22 $\pm$ 0.08 \\ \bottomrule
\end{tabular}
\end{small}
\end{table}

\begin{table}[t]
\centering
\caption{S.U.N.T. rates (\%) for standard ESS-Flow and importance-reweighted samples across tasks.}
\label{tab:quality1multi}
\begin{small}
\begin{tabular}{@{}lcccc@{}}
\toprule
Method & Bulk modulus & Shear modulus & Band gap & Energy above hull \\ \midrule
ESS-Flow & 13.7 & 4.7 & 16.0 & 37.6 \\
Reweighted & 9.5 & 3.3 & 0.9 & 12.3 \\ \bottomrule
\end{tabular}
\end{small}
\end{table}

While the property distributions appear to better match the target, the quality of the reweighted samples as measured by the S.U.N.T. rate in Table \ref{tab:quality1multi} is considerably worse, though this metric may not capture all aspects of sample quality.
The uniqueness rate drops due to resampling from the discrete weighted distribution.
While simple importance reweighting can improve accuracy for moderately concentrated targets, more sophisticated multi-fidelity approaches are needed for sharply concentrated ones, which we leave for future work.

\end{document}